\documentclass[10pt, a4paper]{article}

\usepackage{lrec-coling2024} 
\hypersetup{breaklinks=true, colorlinks=true, citecolor=darkblue, linkcolor=darkblue, urlcolor=darkblue}
\makeatletter
\g@addto@macro{\UrlBreaks}{\UrlOrds}
\makeatother
\usepackage{booktabs}
\usepackage[most]{tcolorbox}
\tcbuselibrary{fitting,skins}
\usepackage{listings}
\newtcblisting{tcbverbatim}{
  fontupper=\tiny,
  blank,
  borderline={1pt}{-2pt},
  nobeforeafter,
  boxsep=1pt,top=0pt,bottom=0pt,
  listing only,
  listing options={
    language=python,
    basicstyle=\scriptsize\ttfamily,
    columns=flexible,
    breaklines,
  }
}

\title{Cross-Lingual Learning vs. Low-Resource Fine-Tuning:
A Case Study with Fact-Checking in Turkish}
         
 \name{Recep Firat Cekinel$^{1, 2}$, Pinar Karagoz$^{1}$, Cagri Coltekin$^{2}$}

 \address{$^{1}$Middle East Technical University \\ Ankara, Turkiye \\ 
          \{rfcekinel, karagoz\}@ceng.metu.edu.tr \\ \\
          $^{2}$University of Tübingen \\ Tübingen, Germany \\ 
          ccoltekin@sfs.uni-tuebingen.de }

\usepackage{tikz}

\abstract{
The rapid spread of misinformation through social media platforms has raised concerns regarding its impact on public opinion. While misinformation is prevalent in other languages, the majority of research in this field has concentrated on the English language.  Hence, there is a scarcity of datasets for other languages, including Turkish. To address this concern, we have introduced the FCTR dataset, consisting of 3238 real-world claims. This dataset spans multiple domains and incorporates evidence collected from three Turkish fact-checking organizations. Additionally, we aim to assess the effectiveness of cross-lingual transfer learning for low-resource languages, with a particular focus on Turkish. We demonstrate in-context learning (zero-shot and few-shot) performance of large language models in this context. The experimental results indicate that the dataset has the potential to advance research in the Turkish language. 
 \\ \newline  }

\begin{document}

\maketitleabstract

\section{Introduction}

Progresses in social networking and social media have not only made information more accessible but have also enabled the rapid spread of false information on these platforms \cite{vosoughi2018spread}. As a result, disseminating fake stories has emerged as a powerful instrument for manipulating public opinion, as observed during the 2016 US Presidential Election and the Brexit referendum \cite{pogue2017stamp, allcott2017social}. Fake news can be described as media content that contains false information with the intent to mislead individuals \cite{shu2017fake, zhou2020survey}. The goal of fake news detection is to evaluate the correctness of statements within the message content.

The traditional method of evaluating the correctness of a claim involves seeking the expertise of specialists who assess the claim by examining the available evidence. For instance, organizations like PolitiFact\footnote{\url{https://www.politifact.com/}} and Snopes\footnote{\url{https://www.snopes.com/fact-check/}} rely on editors to validate the correctness of statements. However, this approach is both time-consuming and expensive. To address this issue, automated methods for fact-checking have emerged, intending to assess the truthfulness of claims while reducing the need for human intervention \cite{oshikawa2020survey}.

Like many other problems in NLP,
the vast majority of available fact-checking resources released are
primarily in English \cite{guo2022survey}.
However, misinformation is not specific to content generated in English.
Automated fact-checking systems are also needed for other languages,
despite having much lower amount of expert annotated fact-checking data.
Besides supervised data availability,
the distribution of languages in pretraining data of state-of-the-art models
also creates a big imbalance between English and other languages.
Since creating large, manually annotated fact-checking data
is a very expensive endeavor,
and finding the amount of unannotated data
in languages other than English to (pre)train large language models
are impractical (if not impossible),
one promising solution is linguistic transfer:
leveraging large datasets in English
and cross-lingual transfer learning methods
to build fact-checking systems for other, low-resource languages.

Cross-lingual learning has been studied in related problems such as hate speech detection \cite{stappen2020cross}, rumor detection \cite{lin2023zero}, abusive language detection \cite{glavavs2020xhate} and malicious activity detection on social media \cite{haider2023detecting}. For fact-checking, \citet{du2021cross} proposed a model that jointly encodes COVID-19-related Chinese and English texts. Additionally, \citet{raja2023fake} employed joint training of English and Dravidian news articles and also applied zero-shot transfer learning by fine-tuning with English data and testing on Dravidian data. 

Our primary aim in this study to
test the viability of cross-lingual transfer learning approaches for fact-checking.
We particularly focus on making use of data in English
for fact-checking in Turkish for the cases
of no or limited data availability.
For this purpose, we collect a fact-checking data set for Turkish,
and perform experiments with transfer learning through 
fine-tuning large language models and utilizing machine translation.
Besides an assessment of the feasibility of transfer learning approaches,
our results also provide some preliminary evidence
for the type of information, knowledge or style,
used in automated fact-checking models.

Our contributions can be summarized as:
\begin{itemize}
    \item Releasing a Turkish fact-checking dataset obtained by crawling three Turkish fact-checking websites.\footnote{\url{https://github.com/firatcekinel/FCTR}}
    \item Assessing the efficiency of transfer learning for low-resource languages, with a specific emphasis on Turkish. 
%
%
    \item Presenting experimental results, comparing zero- and few-shot
      prompt learning and fine-tuning on large language models and underscoring the need to utilize a small amount of native data.
      
\end{itemize}


\section{Related Work}

\paragraph{Datasets.}
In recent years, numerous datasets have emerged for fact-checking and they can be categorized based on how claim statements are obtained.
Some studies that create claim statements by extracting and manipulating content from source documents such as Wikipedia articles can be categorized as artificial claims \cite{thorne2018fever,jiang2020hover,schuster2021get,aly-etal-2021-fact,kim-etal-2023-factkg}.
These studies involve human annotators who systematically generate meaningful claims. 

On the other hand, another approach involves collecting claims by crawling fact-checking websites such as Politifact \cite{vlachos2014fact, wang2017liar} that primarily focuses on political claims and Snopes \cite{hanselowski2019richly} that covers a broader range of topics. Additionally, some studies gather fact-checked claims from the Web \cite{augenstein2019multifc, khan-etal-2022-watclaimcheck}, specifically targeting domains like healthcare \cite{kotonya2020explainable, sarrouti2021evidence}, science \cite{wadden2020fact}, e-commerce \cite{zhang2020answerfact}.
Furthermore, \citet{su2023fake} introduced a hybrid dataset that includes both human-annotated and language model-generated claims.

Fact-checking datasets in languages other than English,
and multilingual datasets are limited in comparison to English.
FakeCovid \cite{shahifakecovid} includes 5182 multilingual news articles related to COVID-19. DANFEVER \cite{norregaard2021danfever}, a Danish fact-checking dataset, comprises 6407 claims generated systematically following the FEVER \cite{thorne2018fever} approach. Similarly, CsFEVER \cite{ullrich2023csfever} features 3097 claims in Czech using a similar methodology. Additionally, CHEF \cite{hu2022chef} contains 10K claims in Chinese. Furthermore, CT-FCC-18 \cite{barron2018overview} contains political fact-checking claims in both English and Arabic, focusing on the 2016 US Election Campaign debates. X-Fact \cite{gupta2021x} comprises 31189 short statements from fact-checking websites across 25 languages. Lastly, Dravidian\textunderscore Fake \cite{raja2023fake} consists of 26K news articles in four Dravidian languages.

The majority of existing datasets have concentrated on textual content for fact-checking. Nevertheless, some claims can benefit from the integration of various modalities, including images, videos and audio.
\citet{resende2019mis} provides video, image, audio and text content from WhatsApp chats to detect the dissemination of misinformation in Portuguese. 
\citet{nakamura2020fakeddit,luo2021newsclippings,abdelnabi2022open, yao2023end, suryavardan2023factify} utilize both visual and textual information for fact-checking. Additionally, MuMiN \cite{nielsen2022mumin} incorporates the social context in the X platform (aka Twitter) and includes 12914 claims in 41 languages. 

To the best of our knowledge,
the only other fact-checking dataset that includes Turkish 
is X-Fact \cite{gupta2021x}
which includes claims and evidence documents in 25 languages.
Besides the differences in size of the corpus,
their Turkish data diverges from ours in a number of ways.
Mainly, our focus in the corpus collection is richer monolingual data,
rather than a large coverage of languages.
The evidence documents in X-fact are through web searches,
rather than crawling directly from the fact-checking site.
Although there is some overlap in our sources,
our data is also more varied in terms of fact-checking sites
and topics of the claims.
We also include short summaries provided in justifications and additional metadata.
The summaries can be valuable for explainability in fact-checking
\cite{atanasova2020generating,kotonya2020explainable,stammbach2020fever,brand2022neural, cekinel2024explaining}.
In addition, a semi-automated method is applied to eliminate duplicate claims that we crawled from different sources.

\paragraph{Methods.}
Automated fact-checking has been studied from data mining \cite{shu2017fake} and natural language processing \cite{oshikawa2020survey, guo2022survey, vladika2023scientific} perspectives. The methods can be classified as content-based and context-based. 

\citet{zhou2020survey} further classify content-based methods
as knowledge-based \cite{ pan2018content, cui2020deterrent}
and style-based \cite{zhou2020fake, perez2018automatic, jin2016novel, jwa2019exbake}.
Both approaches utilize news content to verify the veracity of a statement. While knowledge-based models assess statements by referencing their knowledge base, style-based methods typically prioritize assessing the lexical, syntactic and semantic attributes during verification. 

Similarly, the authors categorized context-based methods
as propagation-based \cite{hartmann2019mapping, zhou2019network} and source-based \cite{sitaula2020credibility}. Both methods aim to capture social context to uncover the spread of information. While propagation-based models leverage interactions among users on social media by enhancing the interaction network with additional details like spreaders and publishers, source-based approaches rely on the credibility of sources which can also be employed to identify bot accounts on social media. 

\citet{kotonya2020explainablesurvey} conducted
a survey of the explainable fact-checking literature
and classified the studies based on explanation generation approaches.These methods include exploiting neural network artifacts
\cite{popat2017truth, popat2018declare, shu2019defend, lu2020gcan, silva2021propagation2vec}, rule-based approaches \cite{szczepanski2021new, gad2019exfakt, ahmadi2020rulehub}, summary generation \cite{atanasova2020generating, kotonya2020explainable, stammbach2020fever, brand2022neural, cekinel2024explaining}, adversarial text generation \cite{thorne2019evaluating, atanasova2020adversarial, dai2022ask}, causal inference  \cite{cheng2021causal, zhang2022causalrd, li2023boosts, xu-etal-2023-counterfactual}, neurosymbolic reasoning \cite{pan-etal-2023-fact, wang2023explainable} and question-answering \cite{ousidhoum-etal-2022-varifocal, yang2022explainable}.

Transfer learning approaches are relatively rare for fact-checking.
One approach in this field focuses on claim matching, aiming to link a claim in one language with its fact-checked counterpart in another language \cite{kazemi2021claim,kazemi2022matching}.
Another approach focuses on out-of-domain generalization,
involving the training of multilingual language models in a cross-lingual context \cite{gupta2021x}.
Besides, cross-lingual evidence retrievers can be employed to retrieve evidence documents in any language corresponding to a claim made in a different language, thereby enhancing the cross-lingual fact-checking capabilities \cite{huang2022concrete}.

\section{Data}

Fact-checking datasets in both Turkish and English, are released by crawling Turkish fact-checking organizations and Snopes for English content. The significant similarity between the fact-checking domains of the Turkish websites and Snopes presents a valuable opportunity for transfer learning. In this study, various experiments are conducted to evaluate the necessity of collecting datasets in low-resource languages versus the effectiveness of transfer learning for these languages. Furthermore, we also conducted topic modeling to explore the latent topics within the datasets in Appendix \ref{sec:topic_modeling} and examined the potential content-based discrepancies between true and fake claims in Appendix \ref{sec:nela}.

\subsection{Dataset for Fact-Checking in Turkish (FCTR)}

We crawled 6787 claims from the three Turkish fact-checking websites:
Teyit, Dogrulukpayi and Dogrula.%
\footnote{\url{https://teyit.org/analiz},\newline \url{https://www.dogrulukpayi.com},\newline \url{https://www.dogrula.org/dogrulamalar}}
All are listed as fact-checking organizations on the Duke Reporters' Lab.%
\footnote{\url{https://reporterslab.org/fact-checking/}}
Dogrulukpayi and Teyit are also members of the International Fact-Checking Network (IFCN) which is a global community of fact-checkers. Our data collection process involved extracting \emph{claim statements}, the corresponding \emph{evidence} presented by the editorial teams, \emph{summaries} providing justifications which are also written by the editors, \emph{veracity labels}, \emph{website URLs} and the \emph{publication dates} of the URLs. 

Claims retrieved from Teyit are summarized using the `findings' section, which provides an overview of the evidence statements. Likewise, when it comes to claims sourced from Dogrula, the summary is derived from the final paragraph within the `evidences' section, encapsulating the key findings. In the case of claims obtained from Dogrulukpayi, the dataset includes a dedicated paragraph following the rating section that encapsulates both the claim and the supporting evidence. This paragraph serves as the summary of these claims. Moreover, unique IDs were assigned to each claim in the dataset. 

\begin{figure}[t]
\begin{center}
\includegraphics[width=7.5cm]{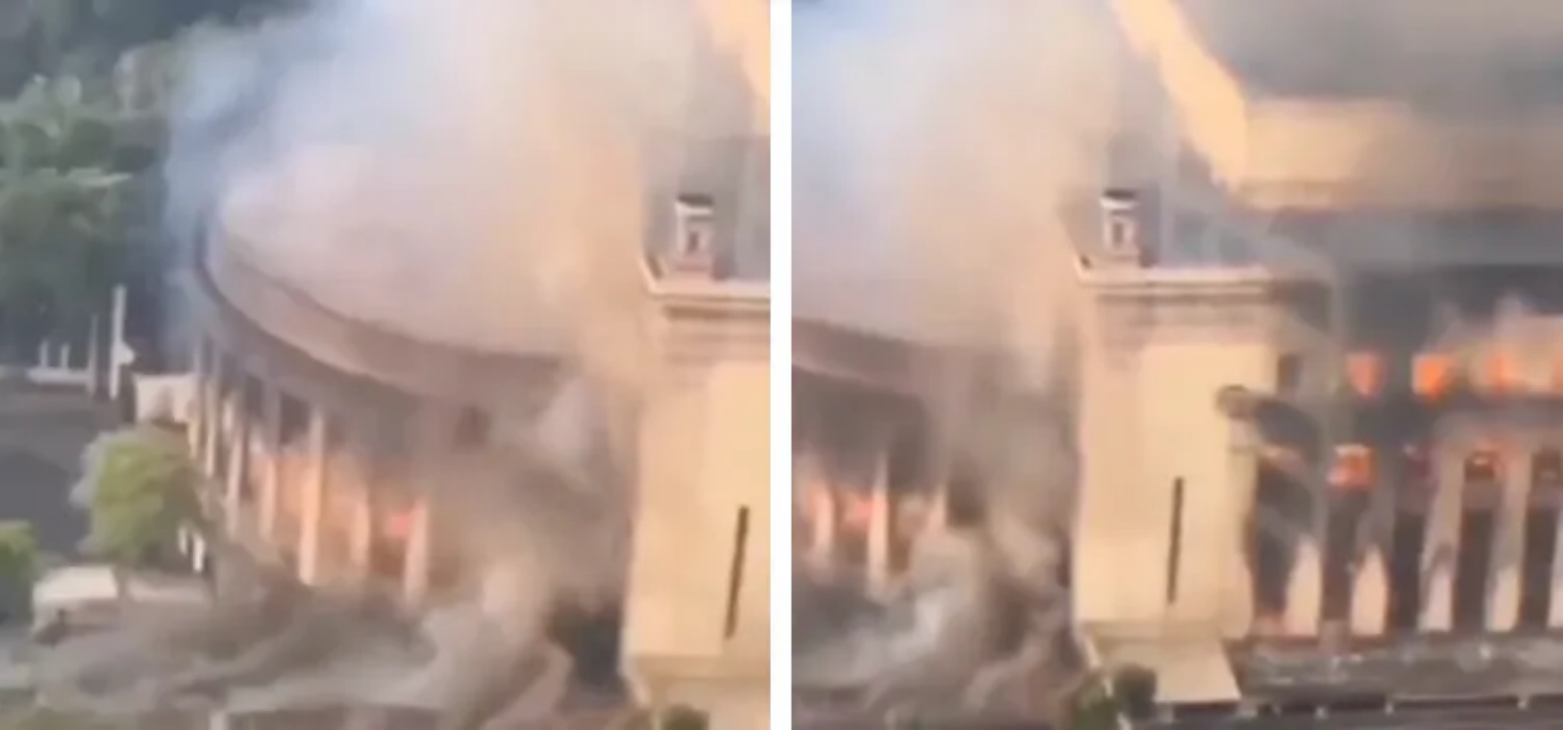}
\caption[Caption for LOF]{A fact-checked claim with multi-modal components \protect\footnotemark}
\label{fig:multimodal_example}
\end{center}
\end{figure}

Claims were also marked as multi-modal if they contained keywords such as `video', `photo' and `image' etc. This classification was made because we recognize that claims featuring such terms require verification not only of their textual content but also of any associated visual or video elements. For example, consider the fact-checked claim presented in Figure \ref{fig:multimodal_example}, which includes an image. In this claim, it was stated that the video shared on social media shows the moments when protesters in France set fire to the Alcazar Library in Marseille during the recent protests. The reviewer who gathered supporting information noted that `According to inverse visual search results, the video is not from Marseille; it's from the Philippines. The building that caught fire is the Manila Central Post Office.' As a result, in order to verify such claims every aspect of evidences should be processed.
Since our focus in this study is linguistic aspects of fact-checking,
we do not make use of claims that require multimodal processing.

\footnotetext{\url{https://teyit.org/analiz/videodaki-yanginin-marsilyadaki-kutuphaneden-oldugu-iddiasi}}

Last but not least, since the claims were collected from three distinct sources, we reviewed the claims to identify candidate duplicate claims. To accomplish this, the BERTScore metric \cite{zhang2019bertscore} was employed that calculates a similarity score by analyzing the contextual embeddings of individual tokens within claim statements. We set the similarity threshold to 0.85 and execute the metric three times in data source pairs. Subsequently, a manual verification process was conducted to confirm whether the outputs from BERTScore indeed corresponded to duplicate claims. 

After the preprocessing step, the dataset contains 3238 claims dating from July 23, 2016 to July 11, 2023. The value counts for each label are presented in Table \ref{tab:fctr-labels}. Furthermore, 742 claims of the final dataset were sourced from Dogrulukpayi, 525 claims were retrieved from Dogrula and 1971 fact-checked claims were gathered from Teyit.

\begin{table}[t]
\centering
  \resizebox{\linewidth}{!}{
  \begin{tabular}{lp{45mm}r}
    \toprule
Label & Sources                      & Counts \\
\midrule
false           & Dogrula, Teyit, Dogrulukpayi & 2780   \\
true            & Dogrula, Teyit, Dogrulukpayi & 203    \\
\midrule
mixed           & Teyit              & 109    \\
partially false & Dogrulukpayi       & 72     \\
unproven        & Teyit              & 37     \\
half true       & Dogrula            & 17     \\
mostly false    & Dogrula  & 14     \\
mostly true     & Dogrula  & 6     \\
\bottomrule
\end{tabular}
}
\caption{Veracity label counts in the FCTR dataset}
\label{tab:fctr-labels}
\end{table}

\begin{figure}[t]
\begin{center}
\includegraphics[width=7.5cm]{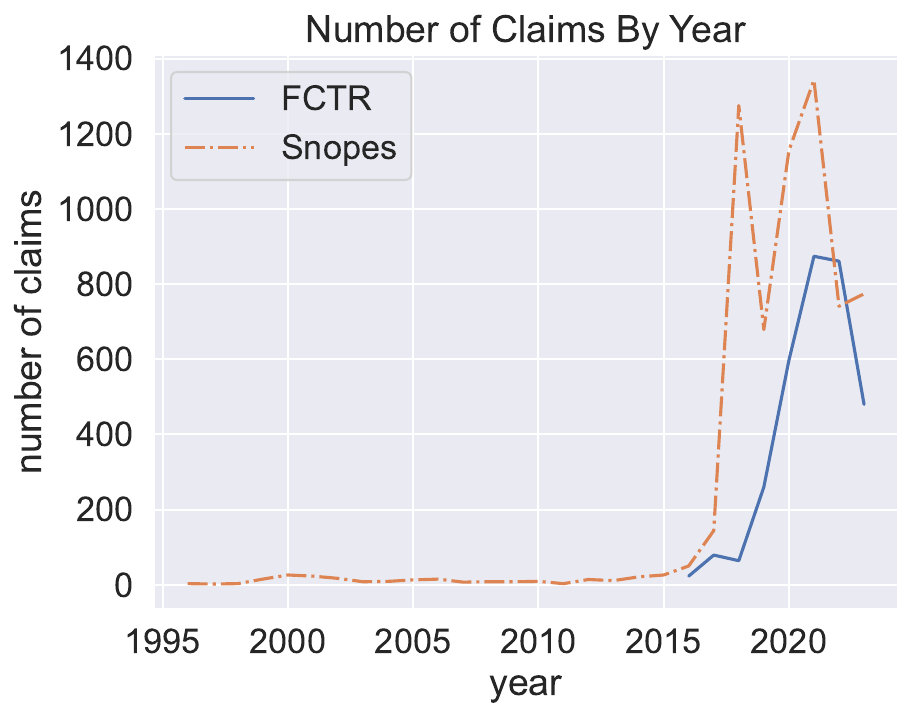}
\caption{Number of claims by year in FCTR and Snopes datasets}
\label{fig:claim-year}
\end{center}
\end{figure}

\subsection{Snopes Dataset}

Snopes is an independent organization committed to fact-checking in English. They employ human reviewers who collect information about claims and write detailed explanations as justifications. It covers a broad range of topics, including politics, health, science, popular culture, etc. We collected claims along with their metadata including the justifications written by human annotators, veracity labels, website URLs and publication dates. We collected 6402 claims ranging from November 24, 1996 to August 17, 2023 and the label distribution is shown in Table \ref{tab:snopes-labels}. Even though Snopes covers a significantly wider date range than the FCTR, the majority of claims are verified within the period from 2015 to 2023 as illustrated in Figure \ref{fig:claim-year}. 

\begin{table}[t]
\centering
\scalebox{0.85}{
\begin{tabular}{lr}
  \toprule
Veracity Labels      & Counts \\
\midrule
false                & 2270   \\
true                 & 1467   \\
mixture              & 588    \\
miscaptioned         & 375    \\
unproven             & 284    \\
labeled satire       & 283    \\
correct attribution  & 247    \\
mostly false         & 237    \\
mostly true          & 198    \\
other               & 453     \\
\bottomrule
\end{tabular}
}
\caption[Caption for LOF]{Veracity label counts in the Snopes dataset\protect\footnotemark} 
\label{tab:snopes-labels}
\end{table}

\footnotetext{`other' encompasses the following labels: scam, outdated, misattributed, originated as satire, legend, research in progress, fake, recall, unfounded, legit}

To the best of our knowledge, Snopes corpus was also crawled by \citet{hanselowski2019richly, augenstein2019multifc}. The reason why we re-collected the Snopes claims is that the previous corpus were released in 2019 but our FCTR corpus is up-to-date. Since we aim to evaluate the effectiveness of cross-lingual transfer learning and considering the potential overlap in fact-checking similar claims across both languages, we gathered the recent fact-checked claims in both English and Turkish.

\section{Method}

\paragraph{Model.}
In this study, we fine-tuned the LLaMA-2 \cite{touvron2023llama} model for the veracity prediction task. Llama-2 is an open-source, auto-regressive transformer-based language model that was released by the Meta AI team. It has three variants, with parameter sizes of 7 billion, 13 billion, and 70 billion. Our main rationale for utilizing Llama-2 is that it has a very large and almost up-to-date knowledge base. To be more specific, the pretraining data includes information up to September 2022, while the fine-tuning data is up to June 2023.
\hfill \break
State-of-the-art language models comprise billions of parameters, demanding large GPU memory resources during fine-tuning for downstream tasks. Additionally, the deployment of such models in real-time applications has become increasingly impractical. Therefore, we adopted parameter-efficient fine-tuning and quantization to make the Llama-2 model fit within our GPU memory constraints without sacrificing information. First, LoRA \cite{hu2021lora} introduces a small number of additional parameters and updates their weights while keeping the original parameters frozen. Similarly, QLora \cite{dettmers2023qlora} employs quantization to the frozen parameters to increase memory efficiency without a significant trade-off. 

\paragraph{Instruction Prompting.}
Instruction tuning is a method that involves additional training of language models using template instruction-output pairs.  It is shown that instruction tuning significantly improves the performance of large language models across a range of tasks \cite{zhang2023instruction}. This is because feeding such tuples to describe the task, allows it to better grasp the domain in question. Additionally, prompting was shown to be an effective way to describe models' reasoning steps by enabling the generation of coherent reasoning chains leading to the desired output \cite{wei2022chain}.

Zero-shot prompting is a method of instructing a language model to generate predictions based on a provided prompt template, without the need for specific examples. During this decision-making process, language models can utilize both the knowledge that they acquired during pretraining and the template prompt. Zero-shot prompting proves particularly useful when you have fine-tuned a language model for a related task but lack labeled data for the specific task at hand. On the other hand, providing one or more examples from the intended task as prompts is referred to as few-shot prompting. By presenting these samples within the prompt, the model gains a better understanding of the desired output and its structure. Therefore, it often leads to superior performance compared to zero-shot prompting.

\section{Experiments and Results}

This section assesses the efficacy of transfer learning in the context of low-resource languages with a specific focus on Turkish. Note that only the best results achieved during the validation experiments for each model are presented.

\subsection{Setup}

The experiments were performed on two distinct datasets: \textit{Snopes} and \textit{FCTR}. Given the highly imbalanced nature of the Turkish fact-checking dataset, we conducted experiments on two variants of \textit{FCTR}, namely \textit{FCTR500} and \textit{FCTR1000}. In the \textit{FCTR500} dataset, all true claims along with 297 randomly sampled false claims were included. Conversely, in the \textit{FCTR1000} dataset, 797 false claims were randomly sampled and combined with 203 true claims. \textit{FCTR500} represents a balanced dataset, while \textit{FCTR1000} serves as its imbalanced counterpart. Other labels were excluded because of their relatively low instance count and the varying labeling conventions within fact-checking communities for ambiguous cases such as partially true and unproven claims. Similarly, when evaluating the language models on the Snopes dataset, we focused specifically on true and false instances. 
In both datasets, we randomly select 80\% of the data for training, 10\% for validation, and 10\% for testing.

\begin{figure*}[t!]
    \begin{tcbverbatim}
### Instruction: Is the following statement "true" or "false"?
### Input: 
A series of photographs show the skeletal remains of the biblical giant Goliath.
### Response:
false \end{tcbverbatim}
    \caption{Prompt template}
    \label{fig:prompt}
\end{figure*}

The SVM model \cite{cortes1995support}
and the multilingual BERT (mBERT) model \cite{devlin-etal-2019-bert}
were both trained on the same datasets with identical
train-dev-test partitions as a baseline.
For the SVM model,
we used sparse word and n-gram features weighted by tf-idf.
The training instances are weighted with inverse class frequency
to counteract the class imbalance,
particularly in the case of \textit{FCTR100} trials.
Similarly, we modified the cross-entropy loss function for the mBERT model.
This adaptation took into account the inverse class ratios,
causing the models to assign a higher penalty
to the errors on the minority class compared to the majority class.

Prompt engineering played a critical role in the experiments. Various prompt formats were evaluated and the best results were achieved using the Alpaca prompt template \cite{alpaca}, which is provided in Figure \ref{fig:prompt}. The LLaMA-2 implementations in the Huggingface's transformers library%
\footnote{\url{https://huggingface.co/meta-llama}} were utilized language models in our transfer learning experiments. Although the LLaMA-2 language model was primarily pretrained on English data, we confirmed its proficiency in Turkish as well. Since it was pretrained on relatively recent data, we preferred LLaMA-2 in our experiments.

In the experiments, we used the SFTTrainer (from trl library) to fine-tune our models. While fine-tuning the LLMs cross entropy loss and Adam optimizer (paged\textunderscore adamw\textunderscore 32bit) with linear scheduler were employed. Additionally, we used a half-precision floating point format (fp16) to accelerate computations. Moreover, we applied parameter-efficient fine-tuning utilizing the QLoRA  \cite{dettmers2023qlora} method to fit the language models to Nvidia Quadro RTX 5000 and Nvidia RTX A6000 GPUs. The configuration included setting the dimension of the low-rank matrices (r) to 16, establishing the scaling factor for the weight matrices (lora\textunderscore alpha) at 64, and specifying a dropout probability of 0.1 for the LoRA layers (lora\textunderscore dropout).

\subsection{Evaluation}

In its prototypical use, fact-checking is very similar
to many retrieval problems.  
We would like to identify a few non-factual texts (e.g., fake news)
among (presumably) many factual documents (legitimate news).
As a result, binary precision, recall and F1 scores
considering non-factual texts as positive instances is
a natural choice for evaluation.
However, the datasets at hand provide an interesting challenge for evaluating
fact-checking models.
Since both classes are obtained from fact-checking organizations,
most claims they care to consider are not factual.%
\footnote{Obtaining claims by other means may be a possible way
to restore the class balance.
However, such an approach also risks introducing spurious correlations
with the veracity label (e.g., topic, style due to collection procedure).}
Hence, the data sets at hand show a reverse class-imbalance
compared to what we expect to observe in real use of such systems.
As a result, for all experiments reported in this paper,
we report F1-macro
and F1-binary scores with respect to the `false' class.
The hyperparameter sweeps are performed to optimize the F1-macro score.

\subsection{Results}



\begin{table}[t!]
\centering
\resizebox{\linewidth}{!}{
\begin{tabular}{llrr}
  \toprule
  Input            & Model     & \multicolumn{1}{c}{F1-macro}  & \multicolumn{1}{c}{F1-binary}\\
 \midrule
claim 10-fold    & SVM                & 0.651 & 0.709 \\
claim            & SVM                & 0.695 & 0.763 \\
claim            & mBERT              & 0.705 & 0.802 \\
claim            & LLaMA-7B           & 0.766 & 0.838 \\
claim            & LLaMA-13B          & 0.814 & 0.866 \\
claim            & LLaMA-70B           & \textbf{0.826} & \textbf{0.890}\\
 \bottomrule
\end{tabular}
}
\caption{Veracity prediction on the Snopes data}
\label{tab:snopes_results}
\end{table}

\paragraph{Snopes Results.}
First of all, we conducted fine-tuning of the LLaMA and baseline models using the Snopes dataset. In all trials, input consisted solely of claim statements, without the inclusion of any supporting evidence. The results are summarized in Table \ref{tab:snopes_results}. According to the results, the LLaMA-2 model with 70 billion parameters exhibited the best performance compared to other models. Since no supporting evidence was provided, the models were expected to rely on stylistic features for their predictions. It is noteworthy that the SVM models learned purely from stylistic features. Nevertheless, a substantial performance gap exists between the SVM and the LLaMA-2 models. This margin could be attributed to the pretrained knowledge embedded in LLaMA-2 models. Moreover, the larger LLaMA-2 models outperformed LLaMA-7B, suggesting that LLaMA-13B and LLaMA-70B leverage their knowledge better than their smaller variant.

\paragraph{FCTR Results.}
Table \ref{tab:fctr500_ftresults} and Table \ref{tab:fctr1000_ftresults} present the fine-tuning results on the \textit{FCTR500} and \textit{FCTR1000} datasets respectively. According to the findings, when using only the claim statement as input, the SVM model which bases its predictions solely on stylistic features achieved the highest F1-macro score on the \textit{FCTR500} and \textit{FCTR1000} datasets. While evaluating with claim statements only, on \textit{FCTR1000} dataset, we fine-tuned the LLaMA models on the Snopes dataset for two epochs initially and continued fine-tuning on the \textit{FCTR1000} dataset for one epoch to achieve the best results. Besides, the class weights of the cross entropy loss function of the multilingual BERT model were adjusted according to the class proportions inversely to get the best result. 

Furthermore, when both the claim statement and the summary (which summarizes the evidence provided by crowd workers) were given as input, the LLaMA-13B model reached a superior 0.89 and 0.828 F1-macro scores on \textit{FCTR500} and \textit{FCTR1000} datasets respectively and 0.923 and 0.947 F1-binary scores respectively. These scores were substantially higher compared to training the model with claims alone. The reason why we incorporated summaries as input was to examine whether this additional information improves the models’ capabilities. Notably, the LLaMA models have limited proficiency in Turkish and we observed poor performance when solely presented with claim statements. 


\begin{table}[t!]
\centering
\resizebox{\linewidth}{!}{
\begin{tabular}{llrr}
  \toprule
  Input             & Model     & \multicolumn{1}{c}{F1-macro} & \multicolumn{1}{c}{F1-binary} \\
\midrule
claim 10-fold     & SVM               & 0.682 &  0.610\\
claim             & SVM               & \textbf{0.714} & 0.709 \\
claim             & mBERT             & 0.653  & 0.750 \\
claim             & LLaMA-7B          & 0.632  & 0.765 \\ 
claim             & LLaMA-13B         & 0.635  & 0.679\\
claim             & LLaMA-70B         & 0.649  & \textbf{0.783} \\
\midrule
+summary    & mBERT         & 0.752 & 0.861 \\
+summary    & LLaMA-13B         & \textbf{0.890} & \textbf{0.923} \\
\bottomrule
\end{tabular}
}
\caption{Fine tuning on the FCTR500 data}
\label{tab:fctr500_ftresults}
\end{table}


\begin{table}[t!]
\centering
\resizebox{7.8cm}{!}{
\begin{tabular}{llrr}
  \toprule
  Input             & Model     & \multicolumn{1}{c}{F1-macro} & \multicolumn{1}{c}{F1-binary} \\
\midrule
claim             & SVM                & \textbf{0.671} & 0.842 \\
claim             & mBERT              & 0.518 & 0.797 \\
claim             & LLaMA-7B           & 0.561 & \textbf{0.864} \\
claim             & LLaMA-13B          & 0.642 & 0.839 \\
\midrule
+summary    & mBERT          & 0.729 & 0.902 \\
+summary     & LLaMA-13B          & \textbf{0.828} & \textbf{0.947}  \\
\bottomrule
\end{tabular}
}
\caption{Fine tuning on the FCTR1000 data}
\label{tab:fctr1000_ftresults}
\end{table}

\begin{table}[t!]
\centering
  \resizebox{\linewidth}{!}{
\begin{tabular}{llrr}
  \toprule
  Model    & Input      & \multicolumn{1}{c}{F1-macro} & \multicolumn{1}{c}{F1-binary} \\
\midrule
LLaMA-7B & 50 claims  & 0.566    & 0.644     \\
LLaMA-7B & 100 claims & 0.570    & 0.716     \\
LLaMA-7B & 200 claims & 0.576    & 0.677     \\
LLaMA-7B & 300 claims & \textbf{0.649}   & \textbf{0.783}     \\
LLaMA-7B & 400 claims & 0.632    & 0.765    \\
\bottomrule
\end{tabular}
}
\caption{Impact of number of inputs on the FCTR500 data}
\label{tab:fctr500-sampling}
\end{table}

\paragraph{Assessing the Impact of Number of Training Instances.}
In this experiment, we examined the influence of varying training data quantities on model performance. We maintained consistency by utilizing the identical test set employed in the previous experiment given in Table \ref{tab:fctr500_ftresults}. Table \ref{tab:fctr500-sampling} illustrates the consequences of manipulating the quantity of training data when employing the LLaMA-7B model. According to the results, as the number of training instances increases, the F1-macro score exhibits gradual improvement. However, when we employed 300 and 400 training instances, the model's performance remained almost constant, with both cases yielding remarkably similar results with only a single instance having a label change in the negative direction. This observation suggests that beyond a certain threshold, additional training instances may not provide substantial performance gains, highlighting the presence of a saturation point in the learning curve.

\subsection{Cross-Lingual Transfer Learning}

Zero-shot learning and few-shot learning can be achieved by providing prompts to large language models. In the zero-shot setting, no specific instances are provided for the given task. Instead, the model makes predictions based solely on the provided instructional prompts and input statements. In contrast, in the K-shot setting, K instances for each class along with their labels are included in the input prompt. This approach enables the model to gain a better understanding of the task's intention and the desired answer format. We evaluated the effectiveness of transfer learning on two distinct datasets: \textit{FCTR500}, which is more balanced, and \textit{FCTR1000}, which is imbalanced. Note that in the experiments, we employed the models that were fine-tuned on the Snopes dataset with the corresponding results provided in Table \ref{tab:snopes_results}.

Moreover, we conducted transfer learning experiments by repeating few-shot settings five times and reported the average scores along with the standard errors. According to Table \ref{tab:fctr500_tlresults} and Table \ref{tab:fctr1000_tlresults}, few-shot learning appears to be beneficial for the LLaMA variants. In other words, providing sample instances within prompts slightly enhanced their performance. However, fine-tuning LLaMA language models with Turkish data resulted in a substantial improvement in the F1-macro score. For instance, on the \textit{FCTR1000} dataset, while few-shot learning achieved the highest F1-macro score of 0.560 (in Table \ref{tab:fctr1000_tlresults}), fine-tuning with Turkish data boosted all LLaMA variants to F1-macro score of 0.642 (in Table \ref{tab:fctr1000_ftresults}).

\begin{table}[t!]
\centering
\resizebox{\linewidth}{!}{
\begin{tabular}{llrr}
  \toprule
  Input             & Model     & \multicolumn{1}{c}{F1-macro} & \multicolumn{1}{c}{F1-binary}\\
\midrule
zero shot & mBERT & 0.550 & 0.667 \\
\midrule
zero shot & LLaMA-7B  & 0.488 $\mp$ 0.026 & 0.577 $\mp$ 0.027 \\
1-shot    & LLaMA-7B  & 0.536 $\mp$ 0.006 & 0.742 $\mp$ 0.009 \\
2-shot    & LLaMA-7B  & 0.545 $\mp$ 0.035 & 0.632 $\mp$ 0.045 \\
3-shot    & LLaMA-7B  & 0.577 $\mp$ 0.011 & 0.642 $\mp$ 0.029 \\
4-shot    & LLaMA-7B  & 0.538 $\mp$ 0.021 & 0.609 $\mp$ 0.024 \\
5-shot    & LLaMA-7B  & 0.533 $\mp$ 0.021 & 0.647 $\mp$ 0.022 \\
\midrule
zero shot & LLaMA-13B & 0.498 $\mp$ 0.014  & 0.699 $\mp$ 0.006 \\
1-shot    & LLaMA-13B & 0.489 $\mp$ 0.026 & 0.683 $\mp$ 0.023 \\
2-shot    & LLaMA-13B & 0.530 $\mp$ 0.028 & 0.689 $\mp$ 0.019 \\
3-shot    & LLaMA-13B & 0.482 $\mp$ 0.022 & 0.670 $\mp$ 0.028 \\
4-shot    & LLaMA-13B & 0.529 $\mp$ 0.036 & 0.638  $\mp$ 0.028 \\
5-shot    & LLaMA-13B & 0.514 $\mp$ 0.013 & 0.632 $\mp$ 0.007 \\
\midrule
zero shot & LLaMA-70B & 0.527 $\mp$ 0.042  & \textbf{0.773} $\mp$ 0.016 \\
1-shot    & LLaMA-70B & 0.507 $\mp$ 0.036 & 0.766 $\mp$ 0.018 \\
2-shot    & LLaMA-70B & 0.539 $\mp$ 0.021 & 0.754 $\mp$ 0.013 \\
3-shot    & LLaMA-70B & 0.492 $\mp$ 0.030 & 0.692 $\mp$ 0.023 \\
4-shot    & LLaMA-70B & 0.542 $\mp$ 0.021 & 0.709 $\mp$ 0.014 \\
5-shot    & LLaMA-70B & \textbf{0.585} $\mp$ 0.017 & 0.709 $\mp$ 0.023 \\
\bottomrule
\end{tabular}
}
\caption{Transfer learning on the FCTR500 data}
\label{tab:fctr500_tlresults}
\end{table}

\begin{table}[t!]
\centering
\resizebox{\linewidth}{!}{
\begin{tabular}{llrr}
  \toprule
  Input             & Model     & \multicolumn{1}{c}{F1-macro} & \multicolumn{1}{c}{F1-binary} \\
\midrule
zero shot & mBERT & 0.529 & 0.736 \\
\midrule
zero shot & LLaMA-7B  & 0.479 $\mp$ 0.019 & 0.647 $\mp$ 0.018 \\
1-shot    & LLaMA-7B  & 0.501 $\mp$ 0.017 & 0.857 $\mp$ 0.013 \\
2-shot    & LLaMA-7B  & 0.518 $\mp$ 0.010 & 0.706 $\mp$ 0.006 \\
3-shot    & LLaMA-7B  & 0.501 $\mp$ 0.010 & 0.691 $\mp$ 0.024 \\
4-shot    & LLaMA-7B  & 0.512 $\mp$ 0.023 & 0.694 $\mp$ 0.024 \\
5-shot    & LLaMA-7B  & 0.502 $\mp$ 0.030 & 0.690 $\mp$ 0.048 \\
\midrule
zero shot & LLaMA-13B & 0.502 $\mp$ 0.011 & 0.803 $\mp$ 0.006 \\
1-shot    & LLaMA-13B & 0.550 $\mp$ 0.016 & 0.811 $\mp$ 0.014 \\
2-shot    & LLaMA-13B & 0.539 $\mp$ 0.033 & 0.788 $\mp$ 0.020 \\
3-shot    & LLaMA-13B & 0.533 $\mp$ 0.017 & 0.763 $\mp$ 0.016 \\
4-shot    & LLaMA-13B & 0.537 $\mp$ 0.010 & 0.758 $\mp$ 0.010 \\
5-shot    & LLaMA-13B & 0.533 $\mp$ 0.029 & 0.737 $\mp$ 0.021 \\
\midrule
zero shot & LLaMA-70B & 0.521 $\mp$ 0.018 & \textbf{0.865} $\mp$ 0.002 \\
1-shot    & LLaMA-70B & 0.528 $\mp$ 0.011 & 0.858 $\mp$ 0.011 \\
2-shot    & LLaMA-70B & \textbf{0.560} $\mp$ 0.033 & 0.841 $\mp$ 0.012 \\
3-shot    & LLaMA-70B & 0.536 $\mp$ 0.023 & 0.806 $\mp$ 0.018 \\
4-shot    & LLaMA-70B & 0.520 $\mp$ 0.019 & 0.808 $\mp$ 0.016 \\
5-shot    & LLaMA-70B & 0.521 $\mp$ 0.018 & 0.778 $\mp$ 0.015 \\
\bottomrule
\end{tabular}
}
\caption{Transfer learning on the FCTR1000 data}
\label{tab:fctr1000_tlresults}
\end{table}

\subsection{Neural Machine Translation}

Neural machine translation is an approach that employs deep learning models to translate a text from a source language to a target language \cite{ranathunga2023neural}. The transformer-based generative large language models are pretrained massively in English. Therefore, their performance in other languages may not be equally impressive. To tackle this challenge, we conducted translations of the Turkish fact-checking dataset into English utilizing the ChatGPT API. Table \ref{tab:chatgpt-results} presents the veracity detection results on the translated data. Note that we employed the models fine-tuned on the Snopes dataset.

The results suggest that employing translated claims led to higher success rates for LLaMA models compared to the few-shot prompting approach. However, the success rate of mBERT was not positively influenced by translation. This phenomenon may be attributed to the differences in pretraining data between LLaMA models and mBERT. To be more specific, the LLaMA models were massively trained on English corpora, while the pretrained data for mBERT might exhibit a more uniform language distribution. 

Additionally, we annotated the test set of \textit{FCTR500} data based on claim statements, marking them as either "local" or "global". Claims that specifically related to Turkiye were marked as "local" claims, while claims with broader implications were labeled as "global". This categorization was done to assess the impact of the LLaMA model's pretrained knowledge on the claim category. We expected that the model would perform better on global claims, given the possibility that it might have pretrained information related to such claims from the web. The results indicate that using the LLaMA-13B model, the average F1-macro for local claims was 0.520 $\mp$ 0.036 while the average F1-macro score for global claims was 0.582 $\mp$ 0.056. However, using the LLaMA-7B model, we obtained the average F1-macro scores of 0.567 $\mp$ 0.017 for local claims and 0.541 $\mp$ 0.015 for global claims. The results imply that the higher F1-macro score for global claims with the larger LLaMA model may be attributed to its pretraining knowledge that should be addressed in further research.

\begin{table}[t!]
\resizebox{\linewidth}{!}{
\begin{tabular}{llrr}
  \toprule
  Dataset    & Model     & \multicolumn{1}{c}{F1-macro} & \multicolumn{1}{c}{F1-binary} \\
\midrule
fctr500  & mBERT     & 0.561    & \textbf{0.789}     \\
fctr500  & LLaMA-7B  & \textbf{0.576} $\mp$ 0.014   & 0.782 $\mp$ 0.007     \\
fctr500  & LLaMA-13B & 0.567 $\mp$ 0.018   & 0.739  $\mp$ 0.013  \\
fctr500  & LLaMA-70B & 0.571 $\mp$ 0.015   & 0.771 $\mp$ 0.007     \\
\midrule
fctr1000 & mBERT     & 0.485    & 0.840     \\
fctr1000 & LLaMA-7B  & 0.524 $\mp$ 0.011   & 0.847 $\mp$ 0.003     \\
fctr1000 & LLaMA-13B & 0.573 $\mp$ 0.013    & 0.879 $\mp$ 0.004     \\
fctr1000 & LLaMA-70B & \textbf{0.581} $\mp$ 0.012  & \textbf{0.883} $\mp$ 0.003\\
\bottomrule
\end{tabular}
}
\caption{Turkish to English machine translation results}
\label{tab:chatgpt-results}
\end{table}

\begin{table}[t!]
\resizebox{\linewidth}{!}{
\begin{tabular}{llrr}
  \toprule
  Dataset    & Model     & \multicolumn{1}{c}{F1-macro} & \multicolumn{1}{c}{F1-binary} \\
\midrule
fctr500  & mBERT     & 0.532    &  \textbf{0.757}   \\
fctr500  & LLaMA-7B  & 0.523 $\mp$ 0.019   & 0.630 $\mp$ 0.023     \\
fctr500  & LLaMA-13B & 0.544 $\mp$ 0.018   & 0.708  $\mp$ 0.006  \\
fctr500  & LLaMA-70B &   \textbf{0.553} $\mp$ 0.025   &  0.725 $\mp$ 0.022   \\
\midrule
fctr1000 & mBERT     &  0.474   &  0.826   \\
fctr1000 & LLaMA-7B  & 0.481 $\mp$ 0.023   & 0.705 $\mp$ 0.020     \\
fctr1000 & LLaMA-13B & 0.552 $\mp$ 0.044    & 0.800 $\mp$ 0.024     \\
fctr1000 & LLaMA-70B &  \textbf{0.556} $\mp$ 0.018  & \textbf{0.832} $\mp$ 0.011\\
\bottomrule
\end{tabular}
}
\caption{English to Turkish machine translation results}
\label{tab:snopestr-results}
\end{table}

Furthermore, we employed Opus-MT's \cite{TiedemannThottingal:EAMT2020} \textit{opus-mt-tc-big-en-tr} model to translate the Snopes dataset into Turkish and subsequently fine-tuned the language models using the translated Snopes' claims. This experiment was conducted to examine the impact of translating an English dataset into a low-resource language, specifically Turkish, on model performance. The fine-tuned models were then evaluated on the test splits of \textit{FCTR500} and \textit{FCTR100} to maintain consistency with the other experiments. According to Table \ref{tab:snopestr-results}, the F1-macro scores slightly decreased compared to the results presented in Table \ref{tab:chatgpt-results} when translating to a low-resource language.

Fine-tuning on translated data involves certain considerations. To be more specific, despite the state-of-the-art machine translation models accurately translating content, it might not be always feasible to maintain all context after translation. Additionally, since the current language models have a better understanding of English, it is an expected outcome that they would exhibit better performance on data translated from Turkish to English. Likewise, the results suggested that collecting native data for low-resource languages (Turkish for this case) is still required to ensure the development of successful models.

\vspace{-2mm}
\section{Discussion}

The main objective of this study is to test the
possibility and the extent of 
making use of a large amount of fact-checking data
and large language models that were heavily pretrained in English
for fact-checking in other languages with much less labeled data,
and much smaller pretraining data for large language models.
We focus on Turkish as a low-resource language for this task.
Although focusing on a single familiar language allows us to 
curate a better fact-checking corpus,
and perform more meaningful error analysis,
our approach is applicable to many languages.
Results are likely to differ based
on typological similarity of the languages in question,
as well other factors like geographical proximity and cultural similarity
of the communities that speak the language.

Our experiments demonstrate some small gains from the high-resource language 
in zero-shot and few-shot settings,
where few-shot learning shows slight improvement over zero-shot.
The results in Table~\ref{tab:fctr500_tlresults} and 
Table~\ref{tab:fctr1000_tlresults} shows a small but 
consistent increase in F1-macro scores when a few examples are included.
The benefit of more few-shot examples is unclear, however.
The same is true for making use of machine translation
from low-resource language to high-resource language.
The test instances translated to English 
labeled by the models trained on English data
clearly better than an uninformed system.
Even a small amount of training data provides better results
than zero- or few-shot approaches.

Another interesting outcome of our results is the success
of small models that rely only on surface cues on the FCTR data.
There are no obvious latent variables (e.g., authors, source websites)
that can identify the veracity label of short claim texts.
This means some relevant information is available on the surface features. 
However, the large language models surpass the simple ones
on English with a large margin (see Table~\ref{tab:snopes_results}).
This may indicate both the help of the linguistic and perhaps factual
information brought by these models.%
\footnote{A potential problem here is these models may have
  the full fact-checking report for the test instances,
  including the clearly stated verdict in their pretraining data.}
However, most probably
the comparatively smaller Turkish data during pretraining
is possibly a factor in low scores of LLaMA with fine-tuning with Turkish
(Tables~\ref{tab:fctr500_ftresults}~and~\ref{tab:fctr1000_ftresults}).

In the majority of the experiments,
only the claim statements were employed as input,
since this is a more realistic scenario as 
individuals typically seek to assess the truthfulness of
a claim before spending time gathering additional information.
We also include evidence statements as input in some experiments,
which show a clear benefit in providing additional information.
However, evidence retrieval is also a challenging problem in fact-checking
(which falls beyond the scope of this study).
A further problem with providing evidence may be
discouraging the model from leveraging its pretrained knowledge
while making decisions.

\vspace{-2mm}
\section{Conclusion}
We present a novel Turkish fact-checking dataset that is collected from three fact-checking resources. It includes 3238 claims with additional metadata from the same resources including evidence and summary of the justifications. The experiments revealed that fine-tuning a large language model on the Turkish dataset yields superior results compared to the zero-shot and few-shot approaches, highlighting the importance of employing datasets for languages with limited resources.



\section{Ethical Considerations and Limitations}

First, we did not process the collected data to ensure anonymization. The dataset encompasses fact-checked claims about public figures including politicians and artists. If any individual mentioned in a claim requests their removal, we can eliminate the associated claims.

Secondly, the data acquisition process adhered to the regulations of the Turkish text and data mining policy. This policy underlies that the datasets can be used exclusively for research purposes. 

Lastly, the Snopes dataset was collected in accordance with the Terms of Use set by Snopes. Therefore, anyone interested in accessing the Snopes dataset must send a request that includes a commitment to use the dataset only for non-commercial purposes.


\section{Acknowledgements}

This research is supported by the Scientific and Technological Research Council of Turkey (TUBITAK, Prog: 2214-A) and the German Academic Exchange Service (DAAD, Prog: 57645447). We would like to thank the anonymous reviewers for their suggestions to improve the study. We also appreciate METU-ROMER and the University of Tübingen for providing the computational resources.

Parts of this research received the support of the EXA4MIND project, funded 
by the European Union´s Horizon Europe Research and Innovation Programme, 
under Grant Agreement N° 101092944. Views and opinions expressed are 
however those of the author(s) only and do not necessarily reflect those 
of the European Union or the European Commission. Neither the European 
Union nor the granting authority can be held responsible for them. 

\section{Bibliographical References}\label{sec:reference}

\bibliographystyle{lrec-coling2024-natbib}
\bibliographystylelanguageresource{lrec-coling2024-natbib}
\bibliography{bibliography}


\appendix
\section{Topic Modeling}
\label{sec:topic_modeling}
\begin{table}[h!]
\centering
\resizebox{0.5\textwidth}{!}{
\begin{tabular}{lrrrrrr}
  \toprule
Dataset        & Topic 1 & Topic 2 & Topic 3 & Topic 4 & Topic 5 & Topic 6 \\
\midrule
FCTR500-train  & 39      & 64      & 105     & 49      & 116     & 27      \\
FCTR500-val    & 8       & 10      & 10      & 9       & 9       & 4       \\
FCTR500-test   & 6       & 9       & 7       & 9       & 15      & 4       \\
\midrule
FCTR1000-train & 73      & 132     & 174     & 130     & 237     & 54      \\
FCTR1000-val   & 9       & 16      & 20      & 18      & 29      & 8       \\
FCTR1000-test  & 12      & 11      & 19      & 21      & 35      & 2       \\
\midrule
FCTR           & 293     & 472     & 524     & 600     & 927     & 167    \\
\bottomrule
\end{tabular}
}
\caption{Topic distribution in the FCTR dataset}
\label{tab:lda-fctr-count}
\end{table}

\begin{table}[h!]
\centering
\resizebox{0.5\textwidth}{!}{
\begin{tabular}{lrrrrrrr}
  \toprule
Dataset      & Topic 1 & Topic 2 & Topic 3 & Topic 4 & Topic 5 & Topic 6 & Topic 7 \\
\midrule
Snopes-train & 206     & 1063    & 386     & 260     & 553     & 327     & 193     \\
Snopes-val   & 26      & 125     & 52      & 27      & 73      & 48      & 23      \\
Snopes-test  & 25      & 124     & 43      & 29      & 75      & 50      & 27     \\
\bottomrule
\end{tabular}
}
\caption{Topic distribution in the Snopes dataset}
\label{tab:lda-snopes-count}
\end{table}

\begin{table}[t!]
\centering
\resizebox{0.5\textwidth}{!}{
\begin{tabular}{lr}
\toprule
\textbf{Topics} & \textbf{Representative Words (transl.)}\\
\midrule
Topic1 & claim, news, person, sharing, information, \\
& account, share, be, child, use \\
Topic2 & photograph, image, account, sharing, share, claim, \\
& video, name, view, use \\
Topic3 & country, Turkiye, year, history, claim, \\
& data, take, be, state, Turkic \\
Topic4 & vaccine, be, virus, claim, work, \\
& human, disease, research, person, impact \\
Topic5 & video, claim, news, be, statement, \\
& sharing, name, history, eat, talk \\
Topic6 & use, product, breeding, water, electricity, \\
& plane, production, year, logo, claim \\
\bottomrule
\end{tabular}
}
\caption{Representative words in \textit{FCTR} dataset}
\label{tab:lda-fctr}
\end{table}

\begin{table}[t!]
\centering
\resizebox{0.5\textwidth}{!}{
\begin{tabular}{lr}
\toprule
\textbf{Topics} & \textbf{Representative Words}\\
\midrule
Topic1 & animal, water, world, report, military, \\
& human, fire, Russian, area, Russia \\
Topic2 & say, people, year, man, know, take, \\
& make, time, go, get \\
Topic3 & image, photograph, show, video, picture, \\
& take, create, appear, film, real \\
Topic4 & Trump, president, Obama, White House, former, \\
& Clinton, President Donald, tweet, Donald Trump, say \\
Topic5 & post, article, news, Facebook, claim, \\
& story, publish, report, page, com \\
Topic6 & state, law, government, report, vote, \\
& bill, United States, federal, election, claim \\
Topic7 & covid, vaccine, health, study, drug, \\
& medical, cause, disease, use, patient \\
\bottomrule
\end{tabular}
}
\caption{Representative words in \textit{Snopes} dataset}
\label{tab:lda-snopes}
\end{table}

Topic modeling is a method for discovering abstract topics in a collection of documents. Latent topics indicate the patterns in the data that can be inferred by the relationships between words that occur in the documents. The output of a topic modeling is a set of abstract topics that are represented by a list of the most representative words in the topic. In our analysis, Latent Dirichlet Allocation (LDA) \cite{blei2003latent} topic modeling is applied to the \textit{Snopes} and \textit{FCTR} datasets to explore the latent patterns using the coherence metric. The coherence score can be used to evaluate the semantic similarity between the words in a topic. 

The topic distributions for each data split are given in Table \ref{tab:lda-fctr-count} and Table \ref{tab:lda-snopes-count} respectively. Even though we did not split the datasets according to the topic ratios, the most dominant and the least frequent topics were preserved in all data splits. For instance, in the \textit{FCTR} dataset, The fifth topic is the most frequent topic in all subsets except \textit{FCTR500-val} in which the given topic is not the most dominant topic by a small margin. Additionally, the sixth topic is the least frequent topic in all splits. 

We utilized lemmatization, employing the Spacy library for English \footnote{\url{https://spacy.io/models/en}} and the Zeyrek library for Turkish \footnote{\url{https://zeyrek.readthedocs.io/en/latest/}}. Table \ref{tab:lda-fctr} and Table \ref{tab:lda-snopes} display the most representative words for each topic. The coherence score for the Turkish dataset within these topics was 0.388, and the perplexity score was -7.699. The average entropy value per document was calculated as 1.50, suggesting a moderate topic distribution level. Similarly, the Snopes dataset achieved a coherence score of 0.450 and a perplexity score of -8.796. Moreover, the average entropy score per document was found to be 1.94 which might indicate that the documents cover multiple related topics without a strong focus on a single one.

\section{NELA Features}
\label{sec:nela}
News Landscape (NELA) features \cite{horne2017just} are manually crafted content-based textual attributes for news veracity detection. The authors divided the features into six classes: style, complexity, bias, affect, moral and event. We applied NELA features to examine the discrepancies of the features for fake and true claims in the FCTR dataset and conducted Tukey's pairwise test \cite{tukey1949comparing} to identify statistically significant differences.

Table \ref{tab:nela-features} presents features that exhibit statistically significant distinctions for \textit{FCTR500} and \textit{FCTR1000}. We computed the NELA features for only claim statements and the results indicate that only a few features demonstrate significant divergence for fake and true claims.

\begin{table}[t!]
\centering
\resizebox{0.5\textwidth}{!}{
\begin{tabular}{llr}
  \toprule
\textbf{Subset}   & \textbf{Feature name}   & \textbf{Adjusted p-value} \\
\midrule
FCTR500  & allcaps              & 0.023            \\
FCTR500  & avg\_wordlen         & 0.018            \\
FCTR500  & coleman\_liau\_index & 0.018            \\
FCTR500  & lix                  & 0.032            \\
\midrule
FCTR1000 & NNP                  & 0.049            \\
FCTR1000 & avg\_wordlen         & 0.048            \\
FCTR1000 & coleman\_liau\_index & 0.045            \\
FCTR1000 & lix                  & 0.048           \\
\bottomrule
\end{tabular}
}
\caption{Statistically significantly different NELA features}
\label{tab:nela-features}
\end{table}

\end{document}